%% file: main.tex
\documentclass[letterpaper, preprint, paper,10pt]{AAS}	% for preprint proceedings
\usepackage{bm}
\usepackage{amsmath}
\usepackage{subfigure}
\usepackage[colorlinks=true, pdfstartview=FitV, linkcolor=black, citecolor= black, urlcolor= black]{hyperref}

\usepackage{siunitx}
\usepackage{cleveref}
\usepackage{algorithm}
\usepackage[noend]{algpseudocode}
\usepackage{booktabs}
\usepackage{overcite}
\usepackage{footnpag}			      	% make footnote symbols restart on each page
\usepackage{graphicx}

\PaperNumber{26-860}

\begin{document}

\title{Semi‑Decentralized Multi‑Spacecraft Collision Avoidance under Communication Constraints}

\author{Grace R. Kim\thanks{PhD Candidate, Aeronautics and Astronautics, Stanford University, 496 Lomita Mall, Stanford, CA 94305},  
Mahdi Al-Husseini\footnotemark[1], Duncan Eddy\thanks{Post-Doctoral Researcher, Aeronautics and Astronautics, Stanford University, 496 Lomita Mall, Stanford, CA 94305}, 
\ and Mykel J. Kochenderfer\thanks{Professor, Aeronautics and Astronautics, Stanford University, 496 Lomita Mall, Stanford, CA 94305}
}

\maketitle{}

\begin{abstract}
Current spacecraft collision-avoidance operations rely on intermittent ground-station contacts, requiring operators to plan with delayed and asynchronously updated information. Consequently, maneuvers must be planned with only intermittent information sharing between operators, raising the question of how much coordination is needed to achieve collision-avoidance performance comparable to centralized planning. As satellite populations continue to grow, understanding this tradeoff is becoming increasingly important for scalable collision avoidance operations. Although decision-theoretic approaches such as partially observable Markov decision processes (POMDPs) capture the sequential and uncertain nature of collision avoidance, existing multiagent extensions typically assume either continuous information sharing or communication models that do not reflect operational ground-station constraints. To explicitly model this intermittent information availability, we formulate the spacecraft-to-spacecraft collision avoidance problem as a semi-decentralized partially observable Markov decision process (SDec-POMDP), where we govern information propagation directly by realistic ground-station visibility windows. Joint maneuver policies are computed using approximate Recursive Small-Step Semi-Decentralized A* (RS-SDA*), following the state-of-the-art A*-based lineage for decentralized multiagent planning. Across a representative suite of conjunction scenarios, semi-decentralized planning recovers near-centralized maneuver quality while requiring 28.5\% fewer synchronization events than continuous coordination. Comparisons with representative rule-based operator heuristics further show that communication-aware planning more consistently achieves the desired operational miss-distance band while minimizing unnecessary trajectory deviation. Together, these results establish a practical planning framework for autonomous collision avoidance under realistic intermittent communication, bridging the gap between idealized centralized coordination and fully decentralized planning execution.
\end{abstract}

\input{Content/01_Intro}

\input{Content/02_probForm}

\input{Content/03_Methodology}
\input{Content/03.5ExperimentalSetup}

\input{Content/04_Results}

\input{Content/05_Conclusions}

\section{Acknowledgment}
This research was supported by the Fannie and John Hertz Foundation and the National Defense Science and Engineering Graduate (NDSEG) Fellowship Program. Specifically, this material is based upon work supported by the Air Force Office of Scientific Research under award number FA9550-25-C-B010 in the amount of currently negotiated tuition and stipend rates.

\bibliographystyle{AAS_publication}   % Number the references.
\bibliography{references}   % Use references.bib to resolve the labels.

\appendix
\input{Content/appendix}
\end{document}

%% file: Content/01_Intro.tex
\section{Introduction}

Low Earth orbit (LEO) has become an increasingly congested and commercially active domain. Over the past decade, the number of active satellites has grown rapidly~\cite{SPENCER2024120, mcdowell_gcat}, driven in part by the expansion of mega-constellations from private industry operators. 
%Proposed systems, such as SpaceX’s orbital data center concept, envision deployments of up to one million satellites~\cite{fcc_da26_113_2026}, representing an increase of approximately two orders of magnitude relative to current satellite populations. 
This growth in the active satellite population is directly increasing the frequency of spacecraft conjunctions, raising collision risk and requiring more frequent collision avoidance operations. In 2023, the United States Space Force 18th Space Defense Squadron (18 SDS) tracked over 600,000 Conjunction Data Messages (CDMs) per day, tripling the daily rate observed in 2020, placing significant operational burden on satellite operators~\cite{ramos2023lessons}. 
Recent FCC space safety reports further highlight this trend, with Starlink executing more than 355,000 collision-risk mitigation maneuvers between June 2025 and May 2026, while Amazon's Project Kuiper also reported thousands of avoidance maneuvers with its constellation expansion~\cite{amazon_leo_status_2026, spacex_gen1_gen2_status_2026}. With multiple active mega-constellations, collision avoidance will increasingly require coordination between multiple independent operators with asynchronous information updates, making scalable communication and planning strategies essential. Collectively, these trends suggest that conjunction management is evolving from a monitoring task into a large-scale, real-time decision-making problem.

Current collision avoidance (CA) operations remain largely ground-driven and operator-specific. Satellite operators receive CDMs from U.S. Space Force or commercial space situational awareness providers, evaluate conjunction risk, and plan candidate maneuvers subject to their own operational constraints~\cite{SORGE2025600}. These decisions are made under uncertainty: state estimates are derived from imperfect observations, covariance estimates are often poorly calibrated, and maneuver outcomes propagate nonlinearly over time~\cite{analyticalGonzalo2021}. Furthermore, this uncertainty is compounded by limited coordination across operators. Decision-making practices for collision-avoidance maneuver planning vary widely among satellite operators, ranging from heuristic thresholds, human-in-the-loop workflows, and proprietary autonomous systems~\cite{SPENCER2024120}. As a result, conjunction mitigation decisions are often made independently, with limited awareness of other agents’ actions, leading to potentially inconsistent or suboptimal outcomes.

Satellite collision avoidance has been studied extensively from operational, analytical, optimization, and decision-theoretic perspectives. Existing operational systems support conjunction screening and risk-based maneuver recommendation~\cite{Braun2016ESACAOps}, while analytical and optimization approaches generate feasible maneuvers using orbit propagation, relative-motion models, and constrained optimization techniques~\cite{analyticalGonzalo2021,GONZALO2020282}. More recently, researchers have explored onboard autonomy through machine learning, reinforcement learning, and Markov decision process formulations for maneuver planning under uncertainty~\cite{Gonzalo2021OnboardCA,kuhlMarkov2025,11357871,harris2019spacecraft}. These developments reflect a broader trend toward increasingly autonomous and data-driven collision avoidance systems~\cite{choumos2024artificial}.
%\cite{Uriot2022CACMLChallenge}
However, the vast majority of existing methods formulate collision avoidance from the perspective of a single spacecraft or operator, rather than explicitly modeling the coupled decision-making process that arises when multiple active spacecraft respond to the same conjunction.

When multiple active spacecraft are involved in a close encounter, conjunction resolution naturally becomes a multi-agent planning problem. The maneuver selected by one operator alters the feasible and safe actions available to the others, leading to a coupled planning problem between actors, rather than an independent optimization. Recent work has explored this interaction through game-theoretic models~\cite{Dolan2023GameTheoreticCollisionAvoidance,blasch2012Orbital} and multi-agent coordination strategies~\cite{Wang2025Automating}, particularly in the context of constellation management and autonomous satellite operations. These approaches demonstrate the value of reasoning jointly over multiple spacecraft. However, comparatively little attention has been given to the operational planning process immediately following conjunction notification, where multiple operators must rapidly construct a coordinated avoidance strategy under uncertainty before maneuver execution.

Collaborative maneuver planning is further complicated by the realities of space-ground communication. Operators receive updated tracking information only during intermittent ground contacts, producing delayed state estimates and limited opportunities for coordination~\cite{EddyOptimal2025,KimScalable2026}. In operational settings, this prevents continuous communication and synchronized state knowledge across all participants. This exposes a gap between fully centralized approaches, which assume complete information and unrestricted coordination, and fully decentralized methods, where agents make decisions independently and do not leverage coordination when communication opportunities are available. In addition, existing work rarely models the information propagation structure explicitly within the decision-making process, such as intermittent ground contacts and the cost of coordination itself. %This motivates a semi-decentralized planning framework that explicitly models when communication opportunities occur during conjunction resolution, enabling operators to compute a joint avoidance strategy while minimizing coordination overhead prior to execution.

To address this gap, we formulate collaborative spacecraft conjunction avoidance as a semi-decentralized partially observable Markov decision process (SDec-POMDP)~\cite{alhusseini2026}, where synchronization opportunities are derived directly from predicted ground-station visibility windows. Rather than assuming either continuous coordination or completely independent planning, the proposed formulation explicitly models when operators regain updated spacecraft information and can synchronize their decisions throughout conjunction resolution. This communication structure naturally captures the asynchronous information flow characteristic of operational collision avoidance while bridging the spectrum between fully centralized and fully decentralized planning. The resulting SDec-POMDP is solved using approximate Recursive Small-Step Semi-Decentralized A* (RS-SDA*)~\cite{AlHusseini2026ijcai}, an approximate heuristic search algorithm for computing near-optimal joint policies under a specified communication schedule, and is tailored here to autonomous spacecraft conjunction avoidance.

The contributions of this work are threefold. First, we introduce an SDec-POMDP formulation for collaborative spacecraft conjunction avoidance that explicitly incorporates intermittent ground-station communication into an otherwise decentralized planning process, allowing planners to synchronize their information during scheduled communication opportunities while making maneuver decisions independently between synchronization events. Second, we demonstrate that communication-aware planning recovers near-centralized maneuver quality with 28.5\% fewer operator information-sharing events than continuous coordination, and characterize which communication opportunities throughout a conjunction are most valuable for coordinated planning. Third, we compare optimized semi-decentralized policies against representative rule-based operator heuristics, demonstrating that communication-aware optimization more consistently achieves the desired operational miss-distance band while minimizing unnecessary mission deviation. Collectively, these results establish semi-decentralized planning as a practical framework for autonomous multi-operator collision avoidance under ground-contact-derived synchronization constraints.

%% file: Content/02_probForm.tex
\section{Problem Formulation}

We consider the problem of a spacecraft conjunction scenario involving two active, maneuverable satellites, $s_1$ and $s_2$, with different operators. The relative state of $s_1$ to $s_2$ is represented in a radial–tangential–normal (RTN) coordinate frame 
\begin{equation}
x = [p_R, p_T, p_N, v_R, v_T, v_N]
\label{eqn:RTN}
\end{equation}
outlining the radial, along-track, cross-track positions and velocities. Each spacecraft is capable of executing impulsive maneuvers by applying a fixed-magnitude $\Delta v$ in the positive or negative along-track direction. This restriction reflects the primary objective of collision avoidance through relative phasing control.
%where maneuvers are intended to modify the timing of the encounter rather than direct spatial repositioning

A conjunction is defined to occur when the relative miss distance $m$ falls below a screening threshold of \qty{5}{\km} during the time horizon leading up to the time of closest approach (TCA), consistent with standard conjunction screening practices from NASA and the US Space Force~\cite{nasa_ca_handbook_2021}. For each conjunction, our formulation adopts a finite 24-hour planning horizon prior to TCA. This choice reflects the rapid growth of state uncertainty in conjunction predictions, which significantly limits the predictive fidelity of CDM-based information beyond a few days prior to closest approach. The resulting decision window focuses planning on the period in which maneuver tradeoffs are most informed and operationally relevant. The 24-hour planning horizon is event-driven, defined as an ordered sequence of decision stages before TCA. Decision stages $k$ occur every two hours, with additional decision stages corresponding to synchronization opportunities induced by predicted ground-station visibility. At every decision stage, whether periodic or synchronization-induced, each spacecraft updates its local belief and selects its next maneuver according to its local policy. Policies are generated offline and uploaded before execution and define a mapping from the agent's current belief of the state to the appropriate maneuver decision. The ground segment serves as the mechanism for reconciling shared information rather than as the decision-making agent.

\emph{Decision Making Regimes. } We define three operational decision-making regimes over this horizon, all initialized from a shared initial belief $b_0$, or a probabilistic estimate of the system state, derived from the CDM. The three planning regimes differ only in which decision stages permit information synchronization. In the centralized setting, information is synchronized at every decision stage $k$, allowing both spacecraft to maintain a common belief throughout execution and providing an upper bound on achievable planning performance. In the decentralized setting, synchronization never occurs after the initial belief $b_0$, so each spacecraft updates only its local belief and executes its local policy independently. In contrast, the semi-decentralized setting lies between these extremes: agents also begin from $b_0$, but alternate between decentralized execution and synchronized information updates. Here, synchronization occurs only at decision stages for which both spacecraft have completed their scheduled ground-station contacts. At these synchronization stages, updated state information becomes available to both spacecraft, allowing their local beliefs to be reconciled before the next action is selected. Between synchronization stages, each spacecraft updates its local belief using only locally available information and executes its local policy independently.

% Between ground-station contact opportunities, agents execute decentralized open-loop policies independently; at synchronization opportunities, agents exchange information, reconcile their information state, and continue executing coordinated policies.

\begin{figure}
    \centering
    \includegraphics[width=0.8\linewidth]{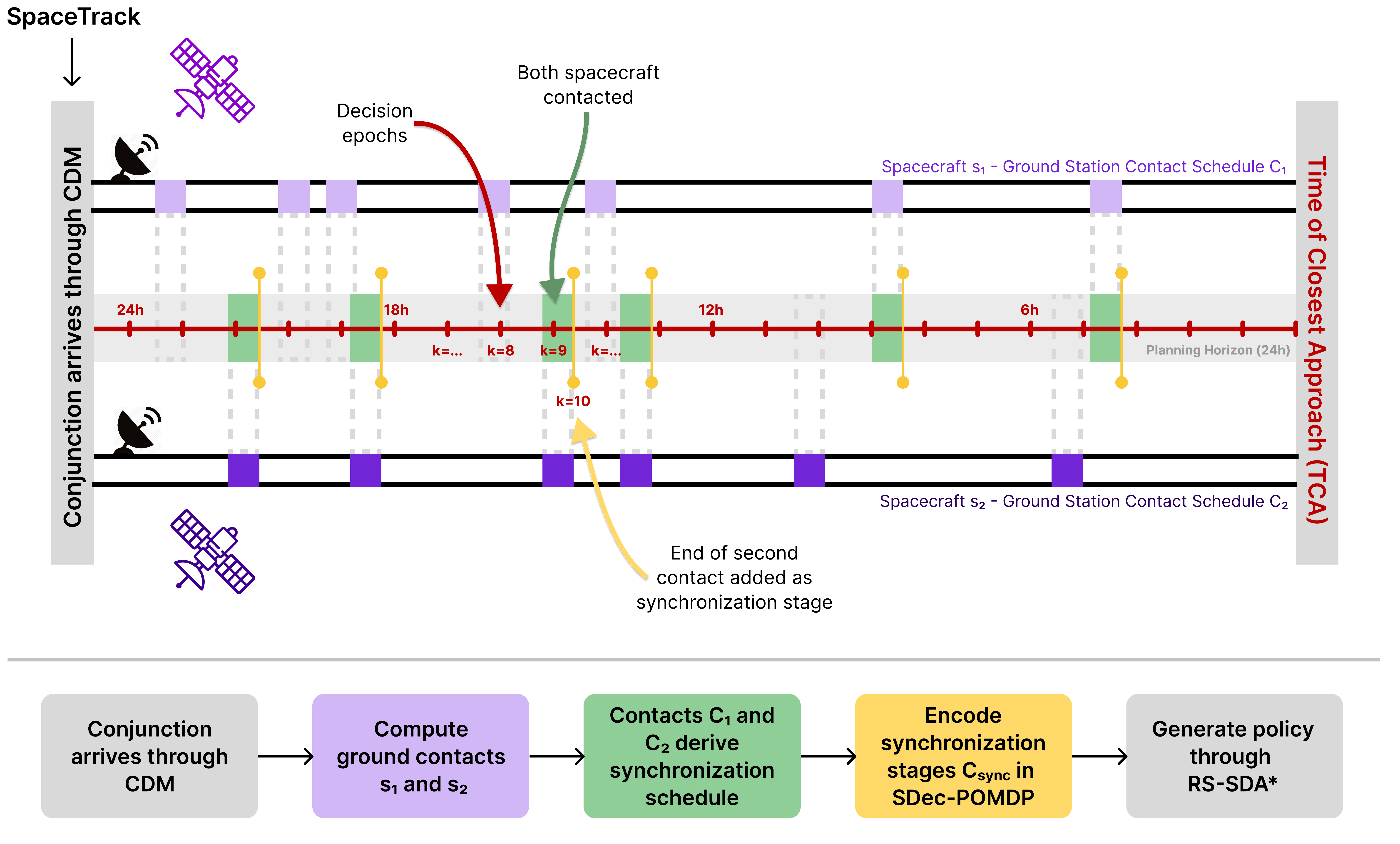}
    \caption{Ground-station-derived synchronization schedule. Individual spacecraft contact windows determine when each vehicle contributes information to the ground-side shared-information state. After both spacecraft have established contact, a synchronization stage is introduced in which the reconciled information is made available to both agents before decentralized execution resumes.}
    \label{fig:groundStationSync}
\end{figure}

\subsection{Ground-Station-Derived Synchronization Schedule}

The remaining question is therefore which decision stages permit information synchronization in the semi-decentralized setting. In our semi-decentralized formulation, information exchange is determined entirely by predicted ground-station visibility, as in practice, spacecraft cannot be assumed to remain in continuous communication with their terrestrial operators. Although some telecommunications constellations maintain near-continuous connectivity, Earth-observation, scientific, and other missions often communicate with their operators only during discrete contact windows. We assume that the spacecraft do not communicate directly with one another. Instead, updated state estimates, conjunction information, and maneuver execution data must first be transmitted to the ground and shared with the other operator through existing ground-segment coordination channels, such as Space-Track.org or operator websites. This dissemination is modeled as an idealized information-sharing mechanism rather than representing the detailed operational workflow. At each synchronization stage, we abstract the complete downlink, ground-side dissemination, and return availability of the reconciled information to both agents as a single information-sharing event, without separately modeling the associated transmission delays. Because the spacecraft generally encounter different contact windows, these updates become available asynchronously. We therefore use predicted ground-station visibility to determine the subset of decision stages at which information synchronization is permitted. %Because conjunctions with different operators will generally experience different contact opportunities throughout the planning horizon, the corresponding operators receive updated information asynchronously. Consequently, the ability to coordinate maneuver decisions is constrained not only by communication itself, but by the timing of ground-station visibility.

%Although the two spacecraft optimize a common collision avoidance objective, they do not maintain synchronized beliefs throughout execution. Each spacecraft maintains its own local belief over the conjunction state using locally available observations, causing their beliefs to diverge between synchronization opportunities. The ground-station communication schedule determines when synchronization may occur, enabling newly available information to be exchanged and the spacecraft to reconcile their beliefs before resuming decentralized execution.

To construct the synchronization schedule, we use the predicted ephemeris of spacecraft $s_i$ and the locations of its supporting ground-station network $\mathcal{L}_i$ to compute the corresponding ground-station contact windows $\mathcal{C}_i$ over the planning horizon. A synchronization stage is introduced once updated spacecraft information from both vehicles has become available through their respective ground-station contacts. Operationally, this corresponds to the later of the two contact opportunities, after which updated state estimates and maneuver execution data from both spacecraft are available for reconciliation. The resulting sequence of synchronization decision stages is denoted by $\mathcal{C}_{\text{sync}}$. Because future ground-station visibility depends only on orbital geometry and known ground-station locations, $\mathcal{C}_{\text{sync}}$ can be computed a priori and remains fixed throughout offline policy generation and subsequent online policy execution. Consequently, synchronization stages are not decision variables, but fixed inputs to the planning problem. \Cref{fig:groundStationSync} illustrates how individual spacecraft contact windows are mapped to the resulting synchronization schedule. 

Between synchronization events, each spacecraft independently updates its local belief using its own observations, maneuver history, and the absence of newly shared information. Because no information is exchanged during this period, the two local beliefs may gradually diverge. At each synchronization event, the operators share updated state information, allowing the spacecraft to reconcile their local beliefs before resuming policy execution. The planner explicitly reasons over this future information structure, allowing maneuver decisions made prior to synchronization to account for the additional information that will become available afterward.
% ^^ sentence mahdi and duncan liked

With this formulation, the communication structure is prescribed by predicted ground-station visibility rather than optimized during planning. The planner therefore optimizes maneuver decisions under a known schedule of future information availability, explicitly reasoning about how upcoming synchronization opportunities influence current maneuver decisions. We now formulate this communication-constrained collision avoidance problem as a semi-decentralized partially observable Markov decision process, where $\mathcal{C}_{\text{sync}}$ defines the communication structure of the multi-agent system.

% The resulting synchronization schedule defines the information-sharing structure used by the semi- decentralized planner. In the centralized case, synchronization occurs at every decision stage, corresponding to continuous information sharing and synchronized beliefs. In the decentralized case, synchronization never occurs after the initial belief is established, resulting in fully independent execution. The proposed semi-decentralized formulation lies between these extremes, with synchronization occurring only at the communication opportunities derived from realistic ground-station visibility.

\subsection{Multi-Spacecraft Collision Avoidance SDec-POMDP}

A SDec-POMDP extends the standard partially observable Markov decision process (POMDP) framework to a multi-agent setting with structured communication constraints. A semi-decentralized partially observable Markov decision process is defined as the tuple
\begin{equation}
(I, \mathcal{S}, \bar{\mathcal{A}}, \bar{\mathcal{O}}, F, T, O, R)
\end{equation}
where $I$ denotes the set of agents, $\mathcal{S}$ is the shared state space, $\bar{\mathcal{A}} = \times_{i \in I} \mathcal{A}_i$ is the joint action space, and $\bar{\mathcal{O}} = \times_{i \in I} \mathcal{O}_i$ is the joint observation space. In the SDec-POMDP, $F$ is a cumulative distribution function governing the likelihood that agents enter a synchronization stage, conditioned on the current state and joint action. %These synchronization stages determine when information may be shared between agents. 
$F$ encodes the synchronization schedule $\mathcal{C}_{\text{sync}}$ defined by the ground-station communication model, outlining when agents may share information to reconcile their beliefs before executing their joint policy. The system evolves according to a transition function $T(s' \mid s, \bar{a})$, which defines the probability of transitioning to state $s' \in \mathcal{S}$ given current state $s \in \mathcal{S}$ and joint action $\bar{a} \in \bar{\mathcal{A}}$. Each agent receives observations according to an observation function $O(\bar{o} \mid s', \bar{a})$, which specifies the probability of joint observation $\bar{o} \in \bar{\mathcal{O}}$ after transitioning to state $s'$ under joint action $\bar{a}$. The reward function $R(s, \bar{a})$ encodes collision avoidance performance, maneuver cost, and mission objectives.

In the multi-spacecraft collision avoidance SDec-POMDP, all communication variants (centralized, semi-decentralized, decentralized) share the same state/action/transition/reward model. They differ only in the information synchronization structure, encoded through synchronization/memory propagation, and the availability of shared observations. %This mathematical formulation is now used to outline the multi-spacecraft collision avoidance problem as a semi-decentralized partially observable Markov decision process.

\emph{State Space.} A straightforward representation of the conjunction state would model the full six-dimensional relative spacecraft state $x$ in the RTN frame from \Cref{eqn:RTN}, where $(p_R,p_T,p_N)$ and $(v_R,v_T,v_N)$ denote the relative position and velocity components between the two spacecraft. However, directly discretizing this six-dimensional state space results in a prohibitively large offline planning problem.

Rather than tracking the full RTN state throughout the encounter, we consider the state of the conjunction event itself, in particular the predicted relative position at the time of closest approach, as the core state information used to inform decision-making.\footnote{A formulation based on probability of collision is also possible, but is considered as future work.} The relative miss distance at TCA is given by
\begin{equation}
m = \sqrt{p_R^2 + p_T^2 + p_N^2}
\label{eqn:miss_distance}
\end{equation}
where $(p_R,p_T,p_N)$ are the components of the predicted relative position at TCA. However, miss distance alone is insufficient as a planning state. Two conjunctions with identical miss distances may require different maneuver decisions depending on whether the spacecraft are predicted to pass ahead of or behind one another and whether previous maneuvers have introduced accumulated orbital drift.

To enable tractable planning while preserving information relevant for decision-making, we exploit the fact that maneuver authority is restricted to fixed-magnitude impulsive burns in the along-track direction. While such burns produce coupled perturbations in all RTN components, their dominant first-order effect is a change in semi-major axis, which induces a differential orbital period between the two spacecraft under the near-circular, small-$\delta v$ regime considered. Over the encounter horizon, this results in secular growth of the relative along-track separation between the two spacecraft in the RTN frame. Radial and cross-track components are weakly affected through higher-order coupling and do not contribute significantly to the leading-order variation in miss distance at TCA for the conjunction scenarios considered. Appendix A provides a detailed validation of this approximation.

\begin{figure}
\centering
\includegraphics[width=0.6\linewidth]{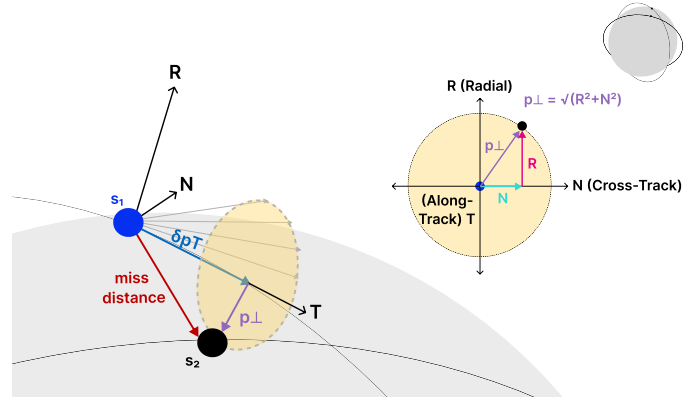}
\caption{Reduction of the six-dimensional RTN relative state to the reduced planning state. The along-track component $\delta p_T$ represents the maneuver-dependent portion of the predicted miss distance at TCA, while the radial and cross-track components are combined into the perpendicular distance $p_\perp=\sqrt{p_R^2+p_N^2}$, which is treated as fixed over the planning horizon. Together, these quantities satisfy $m=\sqrt{\delta p_T^2+p_\perp^2}$ and define the reduced state used by the SDec-POMDP.}
\label{fig:RTNtoDiscretized}
\end{figure}

Preserving the information relevant for decision-making while substantially reducing the state dimensionality, we represent each conjunction using the reduced state
\begin{equation}
s =
(\delta p_T,\,
v_{\text{dev},1},\,
v_{\text{dev},2},\,
k)\label{eqn:reduced_state}
\end{equation}
where $\delta p_T$ denotes the signed along-track separation at TCA, $v_{\text{dev},i}$ represents the accumulated along-track velocity deviation for spacecraft $i$, and $k$ denotes the decision stage. The decision stage is included because the problem is finite-horizon; the remaining time until TCA determines both the available maneuver opportunities and the resulting transition dynamics. The radial and cross-track components of the predicted relative position at TCA are summarized by the scalar
\begin{equation}
p_{\perp}=\sqrt{p_R^2+p_N^2}
\label{eqn:perp_distance}
\end{equation}
which represents the component of the miss distance treated as fixed by the reduced model. Since along-track impulsive maneuvers have negligible influence on $p_{\perp}$ over the planning horizon considered (refer to Appendix A), this quantity is treated as a scenario-specific constant rather than a state variable. Different values of $p_\perp$ therefore represent different ways in which the predicted TCA separation is distributed between the along-track direction and the combined radial and cross-track directions. %However, $p_\perp$ does not by itself characterize the physical encounter geometry, which also depends on the relative velocity vector.

This reduced representation is more informative than representing miss distance alone. In addition to the overall separation magnitude, it preserves (i) the sign of the along-track displacement, distinguishing whether the spacecraft are predicted to pass ahead of or behind one another; (ii) the accumulated orbital drift induced by previous maneuvers through the velocity deviation terms; and (iii) the decomposition of the miss distance into a maneuver-dependent along-track component and a perpendicular component treated as fixed by the reduced model. In particular, including the velocity deviation terms preserves the Markov property of the model, since conjunctions with identical along-track separation but different accumulated drift evolve differently under identical control actions.

Finally, because the offline SDec-POMDP solver requires finite state and transition spaces, each component of the reduced state is discretized into a finite number of bins. This reduced representation enables tractable offline policy computation while retaining the dominant dynamics governing along-track collision avoidance. The discretization procedure is described in the Experimental Setup section.

\emph{Action Space.} Each spacecraft $s_i$ has a discrete action set $\mathcal{A}_i = \{\text{wait}, +\Delta v, -\Delta v\}$, where all maneuvers are modeled as impulsive along-track velocity burns of fixed magnitude of $0.5$~m/s, applied in either the positive or negative along track direction~\cite{Chandramukhi_CAMO_2025,Sales_TrajOptCAM_2013}. For the transition model, these actions are encoded by $a_i \in \{-1, 0, +1\}$, corresponding to $-\Delta v$, wait, and $+\Delta v$, respectively. For a two-spacecraft system, the joint action space is given by the Cartesian product $\bar{\mathcal{A}} = \mathcal{A}_1 \times \mathcal{A}_2$, yielding nine joint actions per stage. Both spacecraft act simultaneously at each decision stage. To maintain a compact finite state space, each spacecraft is limited to one uncompensated maneuver in either direction, while counter-maneuvers remain available to return the velocity-deviation state toward its nominal value; additional velocity-deviation levels could be incorporated at the cost of a larger state space.

\emph{Observation Space.} Each spacecraft receives only a partial observation of the underlying conjunction state. The local observation for spacecraft $i$ is $ o_i = (\delta p_T,\,v_{\text{dev},i})$, where $v_{\text{dev},i}$ is the spacecraft's own velocity deviation, assumed to be observed without error because each operator knows its own maneuver history. The signed along-track separation $\delta p_T$ is shared between agents only during synchronization events.

At synchronization stages, both operators receive a common observation of the current along-track separation, allowing them to reconcile their beliefs regarding the conjunction state before continuing execution. Between synchronization events, no updated observation of $\delta p_T$ is available and the observation instead takes a distinguished null value indicating the absence of new information. Consequently, each agent's observation alphabet consists of the discretized along-track bins together with the null observation and the possible velocity deviation levels.

The joint observation space is given by $\bar{\mathcal{O}}=\mathcal{O}_1\times\mathcal{O}_2$, with the observation function factorizing across agents. The communication architecture considered in this work therefore modifies only the observation model while leaving the underlying state dynamics unchanged. Centralized, semi-decentralized, and decentralized planning differ solely in the stages at which shared observations are available. In the centralized setting, synchronization occurs at every decision stage. In the decentralized setting, synchronization never occurs and each operator observes only its own velocity deviation throughout the planning horizon. The semi-decentralized setting lies between these extremes, with synchronization events induced by predicted ground-station contact opportunities.

\emph{Transition Model.}
The transition function $T(s' \mid s, \bar{a})$ describes how the conjunction state evolves under the joint action $\bar{a} = (a_1, a_2)$. The state includes a predicted along-track separation at time of closest approach, denoted $\delta p_T$. This quantity represents the separation at TCA assuming no further maneuvers and therefore evolves as a forward-predicted metric rather than a propagated orbital state. The stage index advances deterministically as $k \rightarrow k+1$, yielding a finite-horizon, stage-dependent Markov decision process. A maneuver action updates the velocity deviation state $v_{\text{dev},i}$ as
\begin{equation}
v_{\text{dev},i}' = v_{\text{dev},i} + a_i
\qquad
a_i \in \{-1,0,+1\}
\end{equation}
where $v_{\text{dev},i}$ is expressed in units of the fixed maneuver magnitude $\Delta v$ and $a_i$ represents the maneuver increment applied at the current stage. The joint action induces a signed relative maneuver increment $a_{\text{rel}} = a_1 + a_2$. For the predicted along-track separation at TCA, the state variable $\delta p_T$ is updated as
\begin{equation}
\delta p_T' =
\delta p_T +
a_{\text{rel}} \, r(k)
\label{deltaTransition}
\end{equation}
where $r(k)$ is the signed change in along-track separation at TCA produced by an impulsive maneuver applied at stage $k$. This maneuver response is computed by propagating the maneuvered and nominal trajectories from stage $k$ to TCA and evaluating the resulting difference in along-track separation. Because $r(k)$ is defined directly as a displacement at TCA, its dependence on the remaining time to TCA is already incorporated through orbital propagation. The continuous maneuver response is then discretized to construct the transition matrix $T$. Since $\delta p_T$ already reflects previously accumulated displacement, only the change in velocity deviation produced by the current joint action contributes to the update. This ensures that counter-maneuvers stop further divergence without undoing previously accumulated separation.

To model uncertainty, Gaussian noise is added to the deterministic update prior to discretization. The total variance is composed of two independent sources: process uncertainty, which grows with time until TCA $(t_{go}(k))$, and maneuver execution uncertainty, which accounts for independent execution errors in each maneuver commanded at the current stage
\begin{equation}
\sigma^2(k, \bar{a}) =
\sigma_{\text{proc}}^2(k) + \sigma_{\text{man}}^2(k,\bar{a})
\end{equation}
where
\begin{equation}
\sigma_{\text{proc}}(k) =
0.15 \,\frac{\text{km}}{\sqrt{\text{h}}}
\sqrt{t_{\text{go}}(k)}
\qquad
\sigma_{\text{man}}(k,\bar{a}) =
0.02 \, |r(k)| \sqrt{|a_1|+|a_2|}
\end{equation}
and maneuver execution errors associated with simultaneous burns are treated as independent and combined in quadrature. The signed mean maneuver effect is captured by $a_{\text{rel}}$. The resulting stochastic update is
\begin{equation}
\tilde{\delta p}_T' \sim
\mathcal{N}\left(
\delta p_T + a_{\text{rel}} r(k),
\sigma^2(k, \bar{a})
\right)
\end{equation}
This distribution is then projected onto the discretized state space to obtain the transition probabilities.

\begin{figure}[!b]
    \centering
    \includegraphics[width=0.85\linewidth]{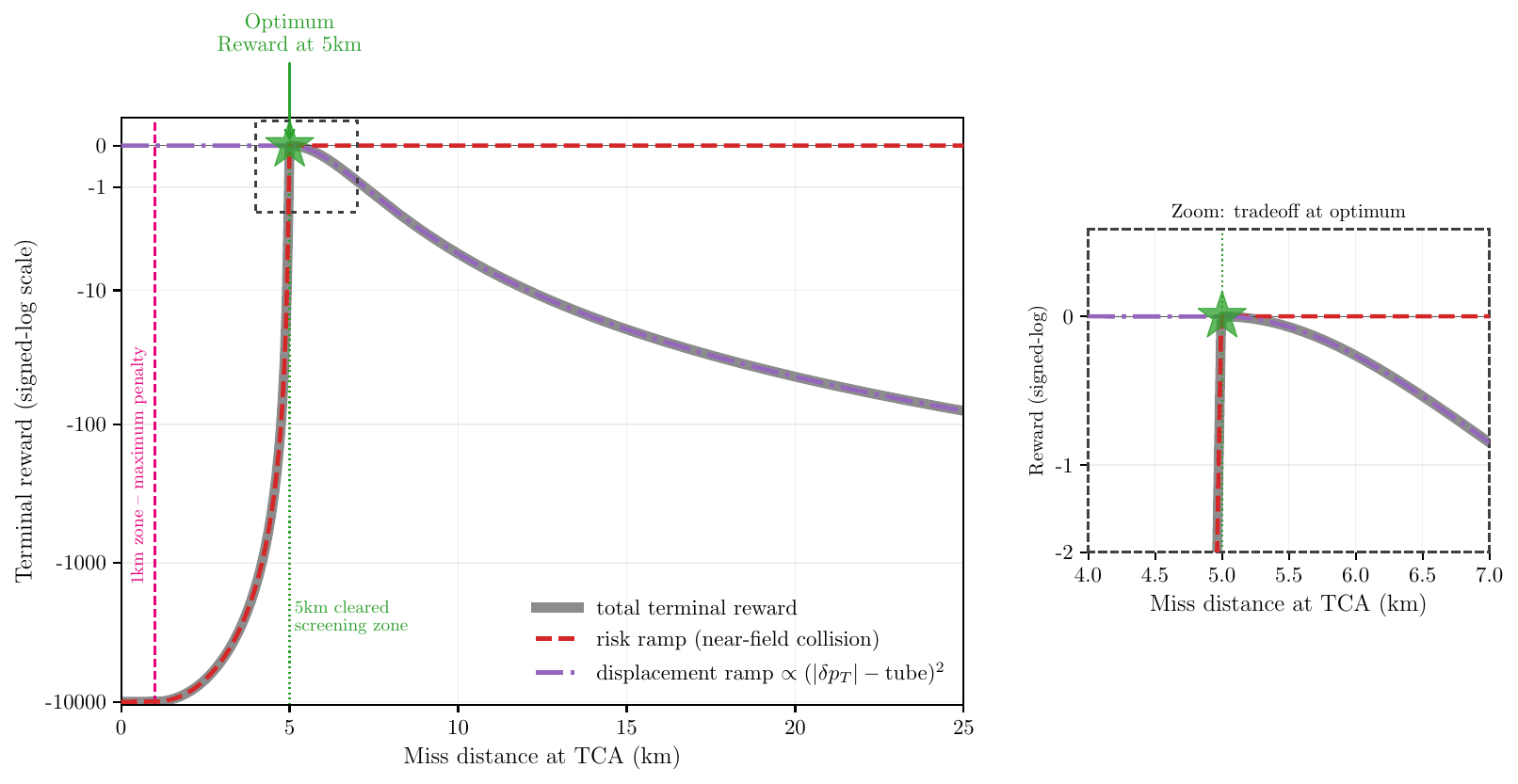}
    \caption{Terminal reward as a function of the miss distance at TCA, plotted on a signed-log scale. A risk ramp (red, floor $-10{,}000$ inside the \qty{1}{\km} collision zone, cleared by \qty{5}{\km}) opposes a quadratic displacement ramp (purple, growing past the \qty{5}{\km} station-keeping tube); their sum (grey) peaks at the $\approx$\qty{5}{\km} ``clear-and-stop" optimum (star). Inset: the $4-$\qty{7}{\km} handoff. The reward favors clearing the threshold with minimal $\Delta v$, not maximizing separation.}
    \label{fig:rewardFunction}
\end{figure}

\emph{Reward Function.}
The reward is designed to capture three competing operational objectives in spacecraft conjunction resolution: (i) ensuring safe separation at time of closest approach, (ii) limiting unnecessary along-track phasing offsets that incur post-encounter recovery cost, and (iii) minimizing the number of impulsive maneuvers. We define the reward as the sum of three components:
\begin{equation}
R(s, \bar{a}) = R_{\text{risk}}(m) + R_{\text{disp}}(\delta p_T) + R_{\text{man}}(\bar{a})
\end{equation}
where $m$ is the miss distance at TCA and $\delta p_T$ is the predicted along-track separation at TCA, both derived from the state representation. This single reward function sums up contributions from both agents into one scalar that the team jointly optimizes. A visualization of the interaction between different reward parameters is provided in Figure~\ref{fig:rewardFunction}.

The risk component penalizes close approaches at TCA. It attains its minimum when the miss distance falls below a collision threshold and smoothly decays to zero outside the screening volume
\begin{equation}
R_{\text{risk}}(m) =
\begin{cases}
-10^4, & m \le 1 \text{ km}, \\
-10^4 \cdot \frac{1 + \cos\left(\pi \frac{m-1}{5-1}\right)}{2}, & 1 < m < 5 \text{ km}, \\
0, & m \ge 5 \text{ km}.
\end{cases}
\end{equation}
to reflect standard conjunction screening practice, where separations beyond the screening volume do not provide additional reward. To avoid solutions that over-separate spacecraft, we penalize excessive along-track offsets
\begin{equation}
R_{\text{disp}}(\delta p_T) =
\begin{cases}
0, & |\delta p_T| \le 5 \text{ km}, \\
-0.2 \cdot (|\delta p_T| - 5)^2, & |\delta p_T| > 5 \text{ km}.
\end{cases}
\end{equation}
to encourage solutions that resolve the conjunction while remaining close to the nominal orbital phasing corridor. The quadratic form reflects increasing cost of orbital restoration for larger deviations. Finally, each impulsive burn incurs a fixed penalty $c_{\text{man}}$
\begin{equation}
R_{\text{man}}(\bar{a}) = -c_{\text{man}} \cdot (|a_1| + |a_2|),
\end{equation}
to discourage excessive actuation, set to $c_{\text{man}}= 2$ for our experiments.

Together, these terms define a structured objective in which collision avoidance dominates, while secondary costs shape operationally realistic behavior. The reward is fully parameterized and can be adapted to different operator preferences without modifying the underlying SDec-POMDP formulation.

%% file: Content/03_Methodology.tex
\section{Planning Methodology}

This section describes the planning and analysis methodology used throughout this work. First, collision avoidance policies are computed using approximate Recursive Small-Step Semi-Decentralized A* (RS-SDA*), which solves the communication-constrained SDec-POMDP formulated in the previous section. We then introduce a synchronization schedule-reduction procedure that identifies a locally minimal subset of ground-station synchronization opportunities needed to recover near-centralized planning performance.

\subsection{Solution Method: Approximate Recursive Small-Step Semi-Decentralized A*}

The proposed problem is solved using approximate Recursive Small-Step Semi-Decentralized A* (RS-SDA*)~\cite{AlHusseini2026ijcai}, an approximate offline planner for SDec-POMDPs. The state, action, and observation spaces are discretized into finite sets to enable offline policy computation prior to execution. % TODO: for a future version include \Cref here
Given the discretized state, action, observation, transition, and reward models together with the synchronization schedule derived from predicted ground-station visibility, approximate RS-SDA* computes a performant joint maneuver policy over the finite planning horizon.

Unlike conventional centralized planners, approximate RS-SDA* explicitly reasons over changing communication availability throughout execution. Between synchronization events, each spacecraft executes its local policy using its current belief, absent of any updated shared conjunction information. When a synchronization event occurs, the planner temporarily reasons over the shared belief state generated from updated information received by both operators, allowing coordinated maneuver decisions before returning to decentralized execution. This naturally matches the communication structure of conjunction resolution, where operators periodically exchange updated spacecraft information but otherwise act independently.

Approximate RS-SDA* incrementally constructs the joint policy through heuristic search over partially specified policies, expanding only the portions of the policy space required to identify high-quality solutions. Computational tractability is achieved through incremental expansion and pruning of suboptimal partial policies~\cite{AlHusseini2026ijcai}. The resulting policy specifies the maneuver each spacecraft should execute as a function of its locally propagated belief and the synchronized information available following communication events. In combination with the SDec-POMDP formulation presented above, the planner explicitly accounts for both orbital uncertainty and the operational communication constraints imposed by intermittent ground-station contacts.

\subsection{Synchronization Schedule Reduction}

Although the synchronization schedule derived from ground-station visibility may contain many communication opportunities, it is unclear which of these synchronization events are actually necessary to recover centralized planning performance. Approximate RS-SDA* assumes that the synchronization schedule is specified a priori as part of the planning model~\cite{AlHusseini2026ijcai}. As a result, the synchronization schedule is treated as a fixed input rather than a decision variable during planning. To better understand the impact of information exchange between operators, we introduce a synchronization schedule reduction methodology. The proposed approach identifies a subset of synchronization events sufficient to recover near-centralized performance relative to the full ground-station schedule. %Although the synchronization schedule derived from orbital visibility may contain many communication opportunities, not every synchronization event necessarily contributes to the resulting maneuver policy. 
Identifying such a reduced set provides insight into where information exchange is most valuable during conjunction resolution.

As described in the problem formulation, the planning horizon consists of a sequence of decision stages $k$. The centralized baseline synchronizes at every decision stage, yielding expected return $R_{\text{cen}}$. In the SDec-POMDP formulation, synchronization is permitted only at the ground-station-derived synchronization stages, denoted by $\mathcal{C}_{\text{sync}}$. Any candidate synchronization schedule is represented by a subset $\mathcal{C} \subseteq \mathcal{C}_{\text{sync}}$, where $\mathcal{C}$ contains a subset of information synchronization events available to the planner. For any candidate schedule $\mathcal{C}$, let $J(\mathcal{C})$ denote the expected return obtained by solving the corresponding SDec-POMDP with synchronization restricted to the events in $\mathcal{C}$. The decentralized baseline therefore corresponds to the empty synchronization schedule,
\begin{equation}
R_{\text{dec}} = J(\emptyset)
\end{equation}
where both $R_{\text{dec}}$ and $R_{\text{cen}}$ define the lower and upper performance bounds, respectively.

%The centralized policy, which synchronizes at every decision stage, provides an upper-performance reference with expected return $R_{cen}$, while the decentralized policy, containing no synchronization after the initial belief, provides the lower-performance bound $R_{dec}$. Each candidate semi-decentralized policy is obtained by solving the same SDec-POMDP while restricting synchronization to a subset $C_{sub}\subseteq \mathcal{C}_{sync}$. All other model components remain unchanged.

Rather than exhaustively evaluating every possible subset of synchronization stages, whose complexity grows exponentially with $|\mathcal{C}_{\text{sync}}|$, we employ a greedy synchronization schedule-reduction procedure to identify a locally minimal synchronization schedule $\mathcal{C}_{\text{sub}}$.  Starting from an initial synchronization schedule $\mathcal{C}_{\text{sync}}$, the algorithm repeatedly attempts to remove a single synchronization event. A removal is accepted only if re-solving the resulting SDec-POMDP produces a policy whose expected return remains within a prescribed tolerance of the centralized solution. This process continues until no remaining synchronization event can be removed without violating the performance criterion. Candidate schedules are evaluated according to their expected return $J(\mathcal{C}_{\text{sub}})$, and are considered equivalent to centralized planning whenever 
\begin{equation}
J(\mathcal{C}_{\text{sub}}) \ge R_{\text{cen}} - \tau.
\end{equation}
where $\tau$ is a small numerical tolerance used to account for floating-point error. Throughout this work, we use $\tau=0.001$, effectively requiring the reduced schedule to reproduce centralized performance.

Because each candidate schedule requires solving a new SDec-POMDP, evaluating many schedules is computationally expensive. To reduce the number of required policy solves, the search is initialized from a burn-centered synchronization schedule rather than the complete schedule. We first compute the centroid of maneuver activity in the centralized solution and construct an initial candidate schedule $\mathcal{C}_{\text{sub}}$ by retaining half of the synchronization opportunities in $\mathcal{C}_{\text{sync}}$, centered about this maneuver epoch. If this burn-centered schedule satisfies the centralized performance criterion, greedy synchronization removal proceeds from this initialization. Otherwise, the algorithm falls back to the complete synchronization schedule $\mathcal{C}_{\text{sync}}$. This warm-start heuristic is motivated by the intuition that information synchronization is most valuable near maneuver execution, reducing the number of expensive SDec-POMDP solves required during the search. \Cref{alg:syncReduction} outlines the approach.

\begin{algorithm}[h]
\caption{Greedy Synchronization Schedule Reduction}
\label{alg:syncReduction}
\begin{algorithmic}[1]
\Require Full schedule $\mathcal{C}_{\text{sync}}$, tolerance $\tau$, expected return of centralized policy $R_{\mathrm{cen}}$
\State Initialize $\mathcal{C}_{\text{sub}}$ using the burn-centered schedule if
$J(\mathcal{C}_{\text{sub}})\ge R_{\text{cen}}-\tau$;
otherwise set
$\mathcal{C}_{\text{sub}}\gets\mathcal{C}_{\text{sync}}$
\Repeat
    \ForAll{$c\in\mathcal{C}_{\text{sub}}$}
        \If{$J(\mathcal{C}_{\text{sub}}\setminus\{c\})\ge
        R_{\text{cen}}-\tau$}
            \State $\mathcal{C}_{\text{sub}}
            \gets
            \mathcal{C}_{\text{sub}}\setminus\{c\}$
            \State \textbf{break}
        \EndIf
    \EndFor
\Until{no synchronization event can be removed}
\State \Return $\mathcal{C}_{\text{sub}}$
\end{algorithmic}
\end{algorithm}

%Starting from the selected initialization, the reduction proceeds iteratively. At each iteration, a single synchronization event is removed, and the resulting SDec-POMDP is re-solved using approximate RS-SDA*. If the reduced schedule remains within the centralized performance tolerance, the removal is accepted; otherwise, the synchronization event is restored. The procedure terminates when no remaining synchronization event can be removed without violating the performance criterion, yielding a locally minimal synchronization schedule for the conjunction. 

% The identified synchronization schedules are subsequently analyzed across all conjunction scenarios to determine common temporal patterns and quantify how many synchronization opportunities are typically required to recover centralized performance.

%% file: Content/03.5ExperimentalSetup.tex
\section{Experimental Setup}

The proposed semi-decentralized collision avoidance framework is evaluated using synthetic conjunction scenarios representative of operational LEO encounters. Each experimental instance consists of a conjunction geometry, a ground-station-derived synchronization schedule, and a scenario-specific SDec-POMDP constructed from high-fidelity orbital propagation. The following subsections describe the ground-station network, conjunction generation procedure, state discretization, and policy evaluation methodology.
\label{sec:Experimental}

\subsection{Ground-Station Network}
\begin{figure}[htbp]
    \centering
    \includegraphics[width=.7\linewidth]{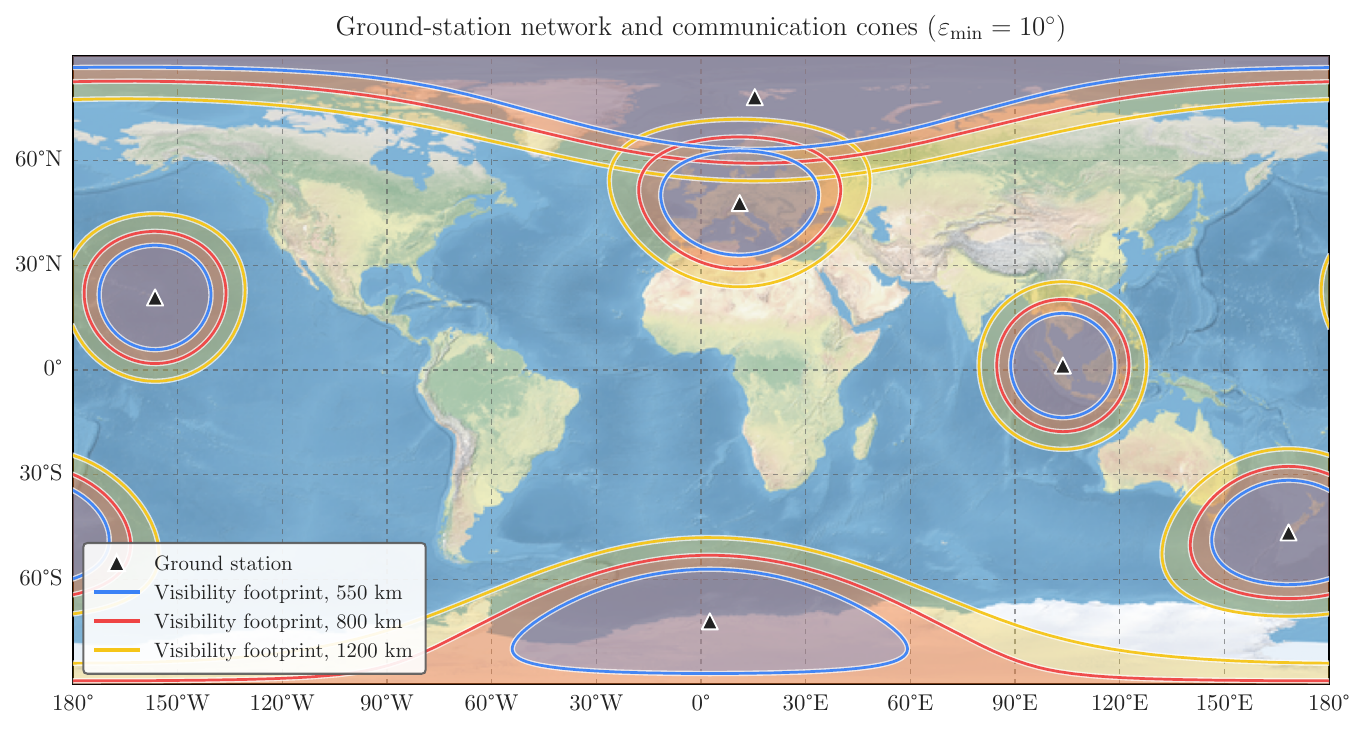}
    \caption{Six-station ground network used throughout the experiments. Communication footprints correspond to a minimum elevation angle of $10^\circ$. Ground-station visibility windows derived from this network determine the synchronization opportunities used by the semi-decentralized planner.}
    \label{fig:GSNetwork}
\end{figure}

Ground-station visibility is modeled using a static six-station subset of the KSAT network~\cite{EddyOptimal2025}, shown in Fig.~\ref{fig:GSNetwork}, consisting of Svalbard, Troll, Hawaii, Singapore, Weilheim, and Awarua. This geographically distributed configuration provides representative global coverage while preserving intermittent communication opportunities between operators. Rather than modeling the complete commercial network, we use a reduced six-station subset, as near-continuous contacts would eliminate meaningful communication constraints by effectively synchronizing the operators at nearly every decision stage. For simplicity, both spacecraft are assumed to share the same six-station ground network. Since ground-station contact windows are computed independently for each spacecraft prior to constructing the synchronization schedule, alternative operator-specific ground-station networks can be incorporated by simply assigning different station sets to each spacecraft.
Ground-station contacts are modeled as deterministic visibility windows that define when each spacecraft can receive updated state estimates and conjunction information. Contact windows are computed using the Brahe library~\cite{eddy2026brahe} with a minimum elevation angle of $10^\circ$. All orbit propagation is performed using a numerical (special perturbations) propagator with a LEO force model including Earth's oblateness ($J_2$), atmospheric drag, solar radiation pressure, and third-body perturbations. Because the two spacecraft generally experience different contact schedules, synchronization events are defined by the later of the two spacecraft contact opportunities, representing the earliest time at which both operators possess updated spacecraft information. This idealized ground-based information exchange defines the synchronization schedule used by the semi-decentralized planner for each conjunction scenario.

\subsection{Conjunction Scenario Generation}
\begin{figure}[t]
  \centering
  \includegraphics[width=.8\textwidth]{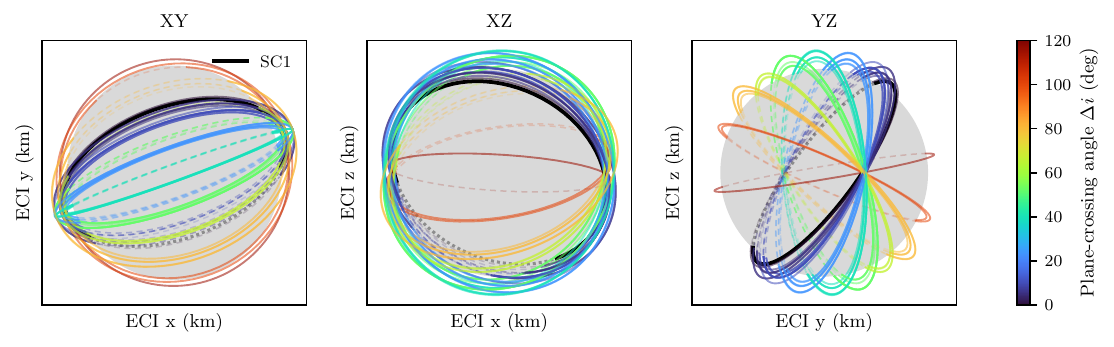}
  \caption{The 52-conjunction evaluation suite, shown as the Earth-centered inertial (ECI) orbit of the secondary spacecraft (SC2) for each scenario. The three panels show the $xy$, $xz$, and $yz$ projections of ECI space. The Earth is shown in gray and the primary spacecraft (SC1) reference orbit in black. Each colored ellipse represents one SC2 orbit, colored by the plane-crossing angle $\Delta i = |i_2-i_1|$ (dark blue: near-coplanar; red: steep crossings). Orbit segments passing behind the Earth are shown as dashed. The scenarios span representative LEO conjunction geometries across the parameter ranges summarized in \Cref{tab:conjunction_parameters}.}
  \label{fig:sweep52}
\end{figure}

Evaluation is performed over 52 representative near-circular LEO conjunction scenarios generated using a parameterized encounter generator as depicted in~\Cref{fig:sweep52}. Candidate conjunctions are restricted to physically realizable near-circular LEO orbits representative of contemporary operational satellites. Although the proposed SDec-POMDP formulation can be adapted to other orbital regimes through appropriate dynamics and communication models, this work's evaluation is limited to LEO, where ground-station contact patterns and encounter dynamics differ substantially from those in MEO, GEO, and highly eccentric orbits. Candidate secondary spacecraft are therefore required to satisfy feasibility constraints on perigee altitude, apogee altitude, and orbital eccentricity before inclusion in the evaluation set.

%The evaluation suite spans conjunction configurations in which miss distances at TCA are set to 1, 2, 5, and 10~km, three reference altitudes (550, 800, and 1200~km), plane-crossing angles up to approximately $115^\circ$, and encounter geometries ranging from nearly head-on to predominantly cross-track. 
Our evaluation suite spans a broad range of conjunction geometries, reference altitudes, and plane-crossing configurations representative of operational LEO encounters. Candidate conjunctions are propagated and verified using the Brahe astrodynamics library~\cite{eddy2026brahe} to ensure the desired closest-approach geometry before inclusion in our test set for evaluation. The resulting dataset provides broad coverage of representative LEO collision geometries while maintaining physically realizable spacecraft orbits. \Cref{tab:conjunction_parameters} provides the swept conjunction scenario parameters. Initial miss distances of $1, 2,$ and \qty{5}{\km} represent conjunctions within or near the screening volume, while the \qty{10}{\km} cases serve as screened-out control scenarios for evaluating whether the planners avoid unnecessary maneuvering.

\begin{table}[t]
\centering
\caption{Conjunction scenario parameters used for experimental evaluation.}
\label{tab:conjunction_parameters}
\begin{tabular}{lr}
\toprule
\textbf{Parameter} & \textbf{Values} \\
\midrule
Number of scenarios & 52 \\
Reference altitude (SC1) & 550, 800, \qty{1200}{\km}\\
Encounter miss distance & 1, 2, 5, \qty{10}{\km} \\
Plane-crossing angle ($\Delta i$) & $0^\circ$--$115^\circ$ \\
Encounter angle & $3^\circ$--$90^\circ$ \\
Orbit type & Near-circular LEO \\\bottomrule
\end{tabular}
\end{table}

\subsection{State Discretization}
\label{sec:discretizationStates}

To build an offline solution via approximate RS-SDA*, the reduced SDec-POMDP state is partitioned into discrete intervals. Each state component is discretized to capture the dominant maneuver-induced state transitions while preserving maneuver-relevant decision boundaries. The resulting discretization is scenario-dependent, since both the planning horizon and the number of available synchronization stages are determined by the conjunction geometry and the ground station contact schedule.

\emph{Along-track separation.} The along-track state $\delta p_T$ is discretized using a signed, non-uniform binning scheme derived from maneuver-induced transition structure. Rather than imposing a uniform grid, bin edges are constructed based on the expected along-track displacement resulting from impulsive maneuvers applied at different time indices within the planning horizon. This is implemented via a deterministic edge-generation procedure that maps time-indexed maneuver effects into state-space boundaries.

To ensure consistency with the reward and evaluation structure, several bin edges are anchored at key operational thresholds, specifically $\{\pm1,\pm4,\pm5,\pm7\}$~km, corresponding to the collision threshold, the desired operational-band boundaries, and the screening and displacement threshold. Additional bin edges are generated by computing the along-track displacement produced by representative impulsive maneuvers executed at each decision epoch. The resulting displacements at TCA are then used as state boundaries, extending the discretization to approximately $\pm50$~km. This aligns bin boundaries with predicted maneuver outcomes, reducing aliasing between states that would otherwise require different control decisions. A central bin spanning $[-1, 1]$ \qty{}{\km} is used to capture the collision-sensitive region. Beyond $\pm 50$ km, absorbing tail bins are introduced to close the state space. Depending on the scenario configuration, this construction yields approximately 20--30 signed bins. This non-uniform discretization better preserves the maneuver-induced transition structure while maintaining a compact state representation.

\emph{Velocity deviation.} Each spacecraft maintains a discrete velocity deviation variable $v_{\mathrm{dev},i} \in \{-1, 0, +1\}$ representing negative, nominal, and positive along-track drift relative to the reference trajectory. This coarse representation is sufficient to distinguish initial avoidance burns from corrective counter-maneuvers while maintaining computational tractability. Repeated maneuvers in the same direction are excluded in the present formulation, although additional velocity-deviation levels could be incorporated by expanding the state space. Across the evaluated scenarios, the resulting optimized policies did not require multiple same-direction maneuvers.

%Higher-resolution discretizations were evaluated but did not materially improve policy structure in the impulsive maneuver regime considered here, while substantially increasing state-space complexity.

\emph{Planning stage.} The planning stage variable $k$ is induced by the ordered sequence of planning events, which includes both regularly spaced decision epochs and, when applicable, ground-station contact opportunities. In the nominal setting, decision epochs occur every two hours over a 24-hour planning horizon. Ground-station contacts computed from orbital visibility introduce additional planning stages when they do not coincide with an existing decision epoch; otherwise, they are merged with the nearest decision stage. In the centralized formulation, agents synchronize at every planning stage. In the decentralized formulation, agents never synchronize and therefore evolve independently throughout the planning horizon. In the semi-decentralized formulation, synchronization is restricted to planning stages associated with ground-station contact opportunities.

\subsection{Model Construction and Policy Evaluation}

Once the discrete state space has been defined, scenario-specific transition, observation, and reward models are constructed before solving each conjunction's SDec-POMDP. The transition, observation, and reward models are generated directly from high-fidelity orbital propagation rather than manually specified transition probabilities. Ground-station visibility determines the synchronization schedule, after which the effect of an impulsive along-track maneuver at each planning stage is computed. These maneuver responses define the transition dynamics and corresponding discretization of the along-track state. Transition probabilities account for process drift and maneuver execution uncertainty through Monte Carlo sampling, while the reward model combines collision-risk penalties, trajectory-deviation costs, and maneuver penalties. The resulting tensors define a scenario-specific SDec-POMDP that is solved offline using approximate RS-SDA*.

Policies are computed offline using the discretized SDec-POMDP model but evaluated through closed-loop propagation in the continuous Brahe astrodynamics environment. Each conjunction is evaluated using 200 Monte Carlo rollouts in the continuous Brahe propagation environment. During each rollout, the continuous spacecraft state is mapped to the corresponding discrete belief state, the policy selects a joint maneuver, and the resulting impulsive burn is applied before propagating the spacecraft to the next planning stage. This evaluation procedure isolates approximation error to the policy synthesis stage while ensuring that reported miss distances and maneuver trajectories reflect the underlying continuous orbital dynamics rather than the discretized planning model.

% Brahe transition matrix generation
% conjunction generation
% ground station network
% discretization parameters
% rollout evaluation
% 52 conjunction scenarios
% solver settings

%% file: Content/04_Results.tex
\section{Results}

Our proposed semi-decentralized framework is evaluated across a representative synthetic suite of conjunction scenarios and compared against fully centralized planning with continuous synchronization and fully decentralized planning without information exchange. Each method's performance is assessed using minimum miss distance at TCA, mission trajectory deviation, and overall planning reward. We first compare the optimized planners against representative heuristic operator strategies to quantify the value of coordinated planning. We then examine whether synchronization restricted to ground-station visibility windows is sufficient to retain centralized planning performance. Finally, the available communication structure is progressively reduced to identify which synchronization opportunities contribute most to coordinated decision-making.

\begin{figure}[t]
    \centering
    \includegraphics[width=.8\linewidth]{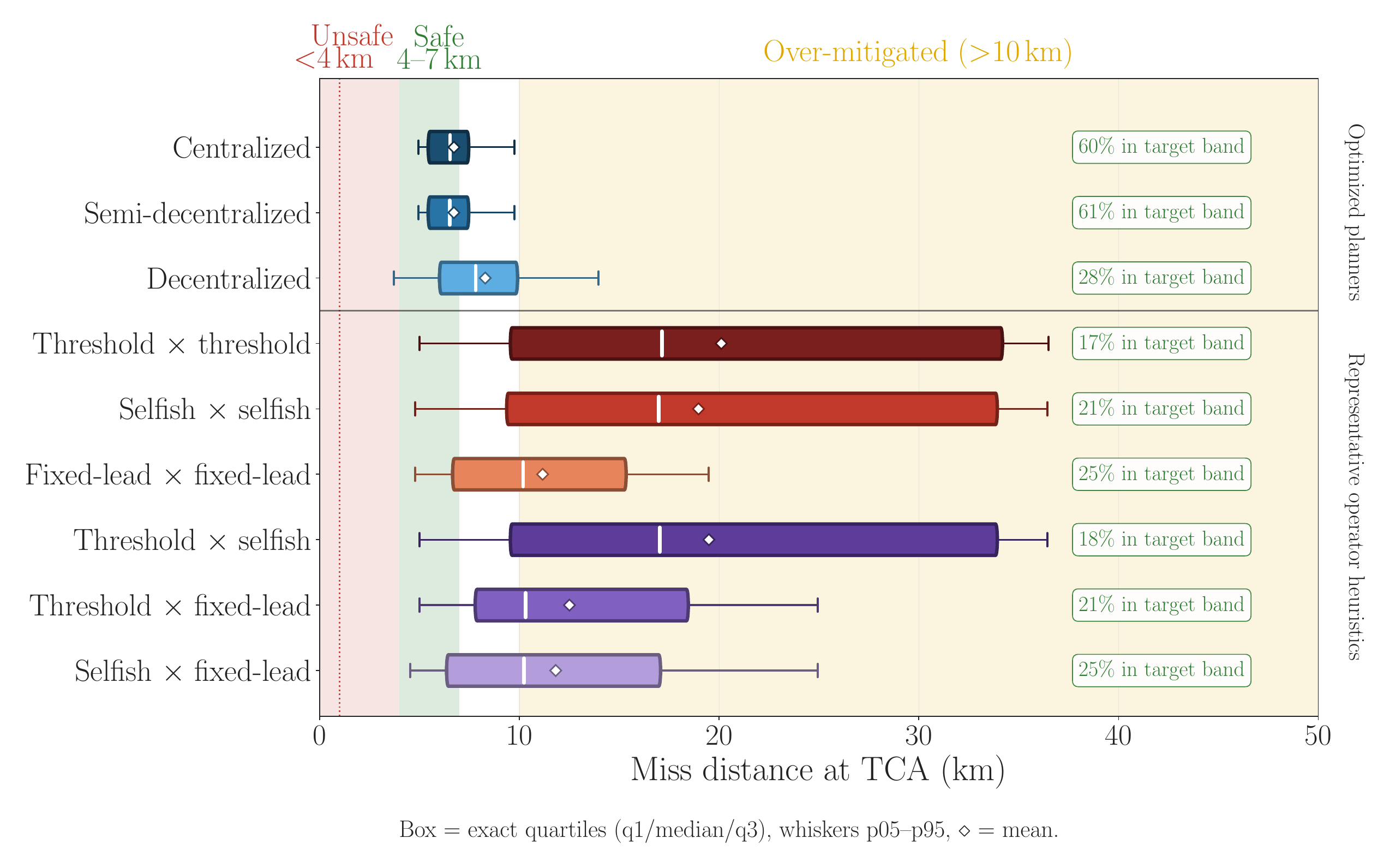}
    \caption{Comparison of optimized planners and representative operator heuristic strategies over Monte Carlo conjunction rollouts. Shaded regions indicate the unsafe conjunction regime ($<$4 km), the desired operational miss-distance band (4–7 km), and increasingly conservative over-mitigation ($>$10 km). Optimized planners consistently target the desired operating region, whereas representative heuristic operators generally satisfy the safety objective by producing substantially larger-than-necessary miss distances. Boxes indicate the interquartile range, whiskers denote the 5th–95th percentiles, diamonds show the mean, and percentages indicate the fraction of rollouts within the target operating band.}
    \label{fig:heuristics}
\end{figure}

\subsection{Comparison with Representative Operator Heuristic Policies}

% While the preceding experiments compare different communication structures within the SDec-POMDP framework, current conjunction operations frequently rely on simpler rule-based operator procedures rather than optimal multi-agent planning. To evaluate the value of optimization, the proposed planners are compared against representative heuristic operator policies that approximate common operational decision philosophies, including threshold-based maneuvering, independent selfish optimization, and fixed-lead maneuver scheduling. Each heuristic independently selects maneuvers using only locally available information and does not optimize a joint policy under uncertainty.
Current conjunction operations frequently rely on rule-based procedures rather than optimized multi-agent planning. We therefore first compare the proposed planners against representative heuristic operator policies that approximate common operational decision philosophies, including threshold-based maneuvering, independent selfish optimization, and fixed-lead maneuver scheduling. The methods are further outlined in Appendix B. Each heuristic selects maneuvers independently using only locally available information and does not optimize a joint policy under uncertainty. For all 52 conjunction scenarios, policies are evaluated using the same high-fidelity truth simulation. Accordingly, all reported final miss distances are determined from propagated spacecraft trajectories rather than from the reduced-order dynamics used during planning.

\Cref{fig:heuristics} compares the resulting miss-distance distributions for optimized planners and representative heuristic operator policies. Centralized and semi-decentralized planning consistently concentrate rollouts within the desired $4$--\qty{7}{\km} operational miss-distance band, with 60\% and 61\% of rollouts terminating within the target region, respectively. This is more than double the performance achieved by the decentralized planner, which reaches the target band in only 28\% of rollouts despite remaining collision-free. The representative heuristic operator policies perform similarly to or worse than the decentralized planner, placing only 17\%--25\% of rollouts within the desired operating region while frequently producing unnecessarily large miss distances that correspond to excessive mission deviation.

These results demonstrate that the primary benefit of the SDec-POMDP framework is not simply maintaining collision safety, but improving maneuver quality under realistic communication constraints. Whereas representative heuristic policies generally satisfy the safety objective by over-mitigating the conjunction, the optimized planners consistently recover solutions that remain close to the desired operational miss-distance band while minimizing unnecessary trajectory deviation. Communication-aware optimization therefore provides benefits beyond collision avoidance alone, producing maneuver strategies that better satisfy both safety and mission-performance objectives than representative rule-based operator policies.

\subsection{Semi-Decentralized Performance Relative to Centralized Planning}

\begin{figure}[!b]
    \centering
    \includegraphics[width=\linewidth]{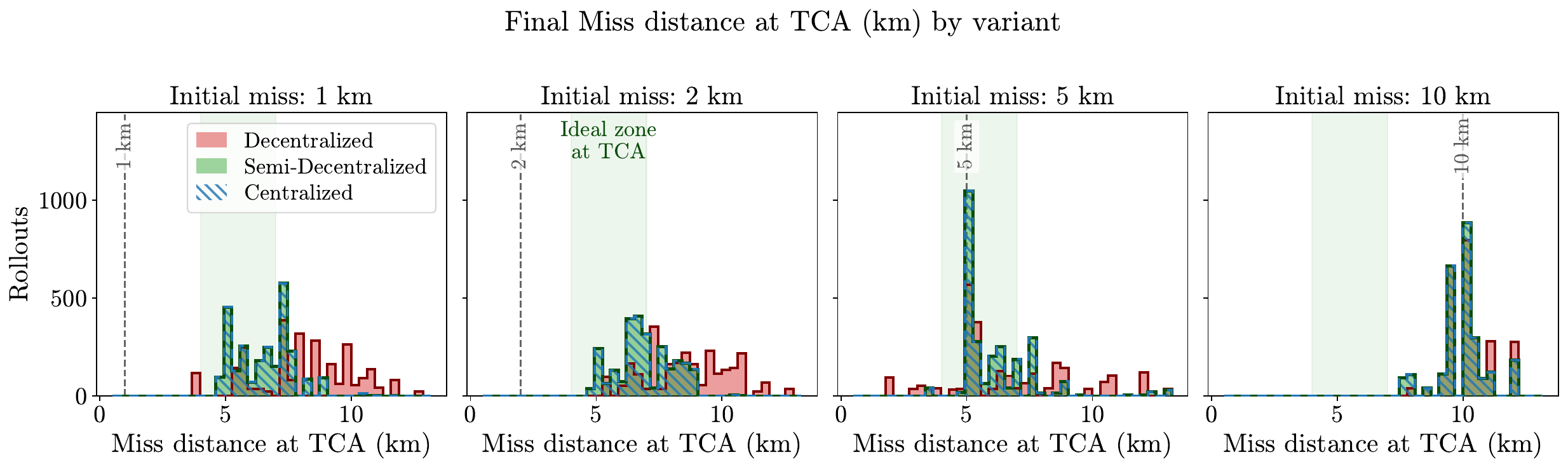}
    \caption{Distribution of final miss distances at TCA across 52 simulated conjunctions for centralized, semi-decentralized, and decentralized planning. Results are shown for varying conjunction scenarios with starting miss distances at TCA set at $1, 2, 5,$ and \qty{10}{\km}. Initial belief of the miss distance is set to a distribution with $\sigma$ = \qty{1.4}{\km} around the true miss distance. Semi-decentralized planning closely matches the centralized solution across the conjunction suite while consistently outperforming fully decentralized planning under realistic ground-station synchronization constraints. The shaded green region denotes the desired miss-distance $4-$\qty{7}{\km} operational band.}
    \label{fig:sweep50init0.5}
\end{figure}

We next evaluate whether synchronization limited to realistic ground-station visibility windows is sufficient to recover the performance of fully centralized planning. Across 52 simulated conjunction scenarios, centralized, semi-decentralized, and fully decentralized planners are evaluated under identical conjunction geometries and maneuver constraints. Both spacecraft use the common ground-station network described in \Cref{fig:GSNetwork}; however, differences in orbital geometry produce distinct contact schedules for each conjunction. Synthetic conjunctions are generated with initial expected miss distances of 1, 2, 5, and \qty{10}{\km} at TCA. Semi-decentralized policies synchronize only when both operators receive updated information following scheduled ground-station contacts, whereas centralized planning assumes continuous shared information throughout execution.

\Cref{fig:sweep50init0.5} demonstrates that semi-decentralized planning closely reproduces the miss-distance distributions achieved by the centralized planner despite synchronization being restricted to realistic ground-station visibility windows. The decentralized planner produces a noticeably broader miss-distance distribution, frequently maneuvering well beyond the desired operational band. 

Across the conjunction suite, centralized planning performs an average of 27.2 synchronization events, whereas the semi-decentralized planner requires only 19.4, representing a 28.5\% reduction in communication. Despite this reduction, both approaches produce identical average planning returns, identical mean propagated miss distances (6.95 km), and indistinguishable maneuver deviations to numerical precision, with zero collisions observed across all evaluated scenarios. By contrast, the fully decentralized planner eliminates synchronization entirely but experiences reduced planning performance, yielding lower average returns and larger propagated miss distances (7.94 km).

Although the fully decentralized planner achieves lower overall performance, no evaluated rollout for any coordination strategy falls below the \qty{1}{\km} collision threshold. Its principal disadvantage is instead maneuver quality, as it more frequently over-mitigates the conjunction and terminates outside the desired operational band, although this conservatism is less pronounced than that of the representative operator-heuristic strategies. This indicates that a jointly generated collision avoidance policy, computed from the initial conjunction assessment and executed without subsequent information updates, remains sufficiently conservative to satisfy the highest-priority operational objective of preventing collision. Once safety is assured, the principal benefit of additional communication shifts from collision avoidance to mission efficiency. Periodic synchronization enables operators to incorporate updated spacecraft state estimates and maneuver execution information as the conjunction evolves, reducing unnecessary conservatism without sacrificing the achieved safety margin. Consequently, semi-decentralized coordination recovers centralized planning performance with substantially fewer synchronization events, providing a practical balance between operational communication constraints and maneuver efficiency. These results establish semi-decentralized planning as an effective compromise between the unrealistic assumption of continuous coordination and the conservatism of completely independent planning.

\begin{figure}[t]
    \centering
    \includegraphics[trim={0cm 0cm 0cm 0.7cm}, clip, width=\linewidth]{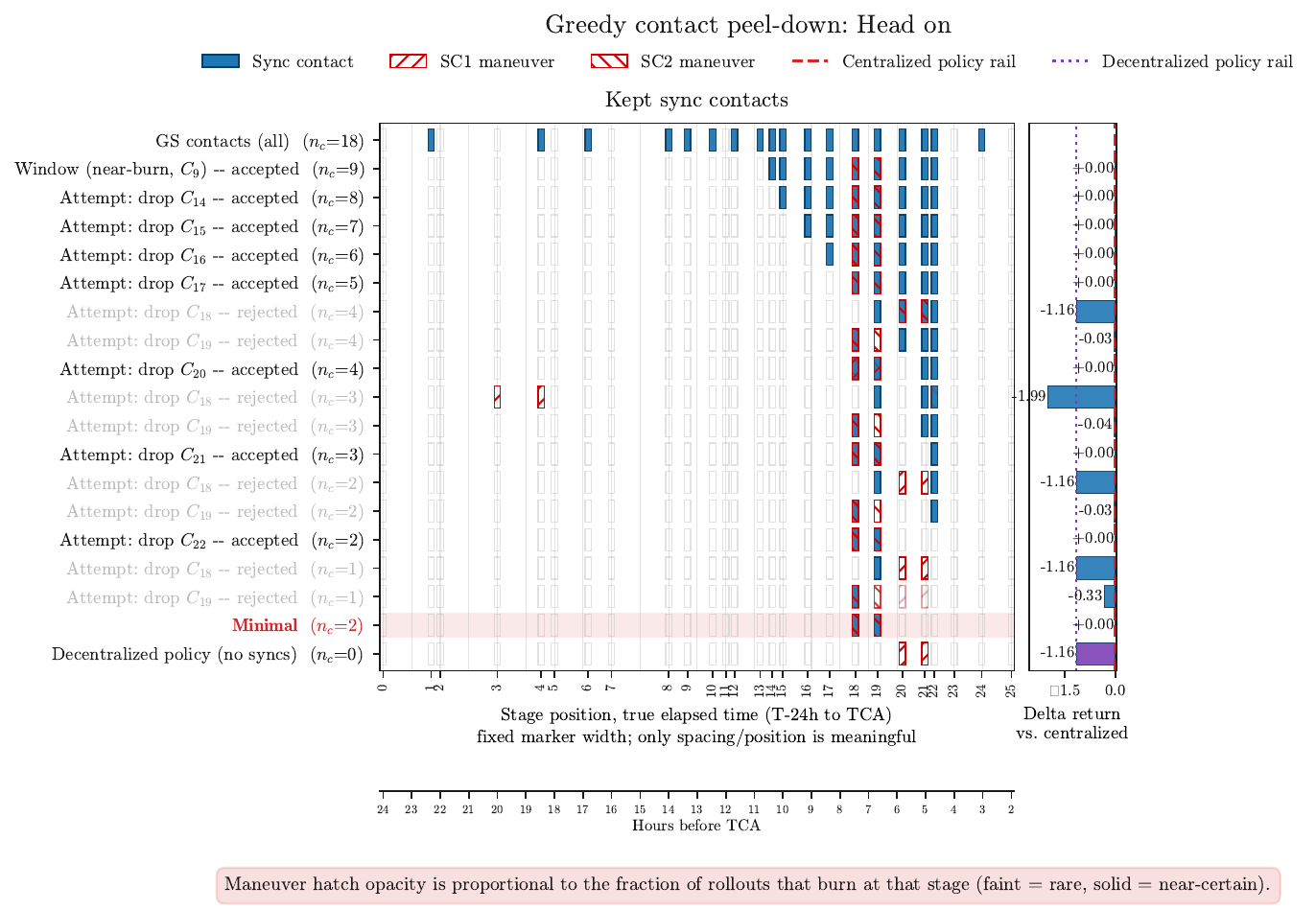}
    \caption{Greedy synchronization-schedule reduction for a representative head-on conjunction. Each row is one evaluated contact subset, with blue cells marking retained synchronization stages. Intermediate rows include both accepted and rejected single-contact removals; the right panel shows the resulting return relative to the centralized solution. Maneuver hatching denotes SC1 and SC2 burn activity, with opacity proportional to maneuver frequency across Monte Carlo rollouts. The bold Minimal row is the locally minimal schedule returned by the greedy search.}
    \label{fig:peelheadon}
\end{figure}

\subsection{Identifying Critical Synchronization Opportunities}

Although semi-decentralized planning substantially reduces synchronization relative to the centralized baseline, communication opportunities remain constrained by ground-station availability and operational considerations. This raises the question of whether every synchronization opportunity contributes equally to coordinated planning performance. To investigate this, two representative conjunction scenarios are examined using the greedy synchronization schedule reduction procedure described in \Cref{alg:syncReduction}. Beginning with the complete ground-station-derived synchronization schedule, synchronization events are iteratively removed while maintaining planning performance within a specified tolerance of the centralized solution. Rather than statistically characterizing the entire conjunction suite, these case studies illustrate how the reduction procedure identifies the communication opportunities that contribute most to coordinated decision making.

\begin{figure}[h]
    \centering
    \includegraphics[trim={0cm 0cm 0cm 0.7cm}, clip, width=\linewidth]{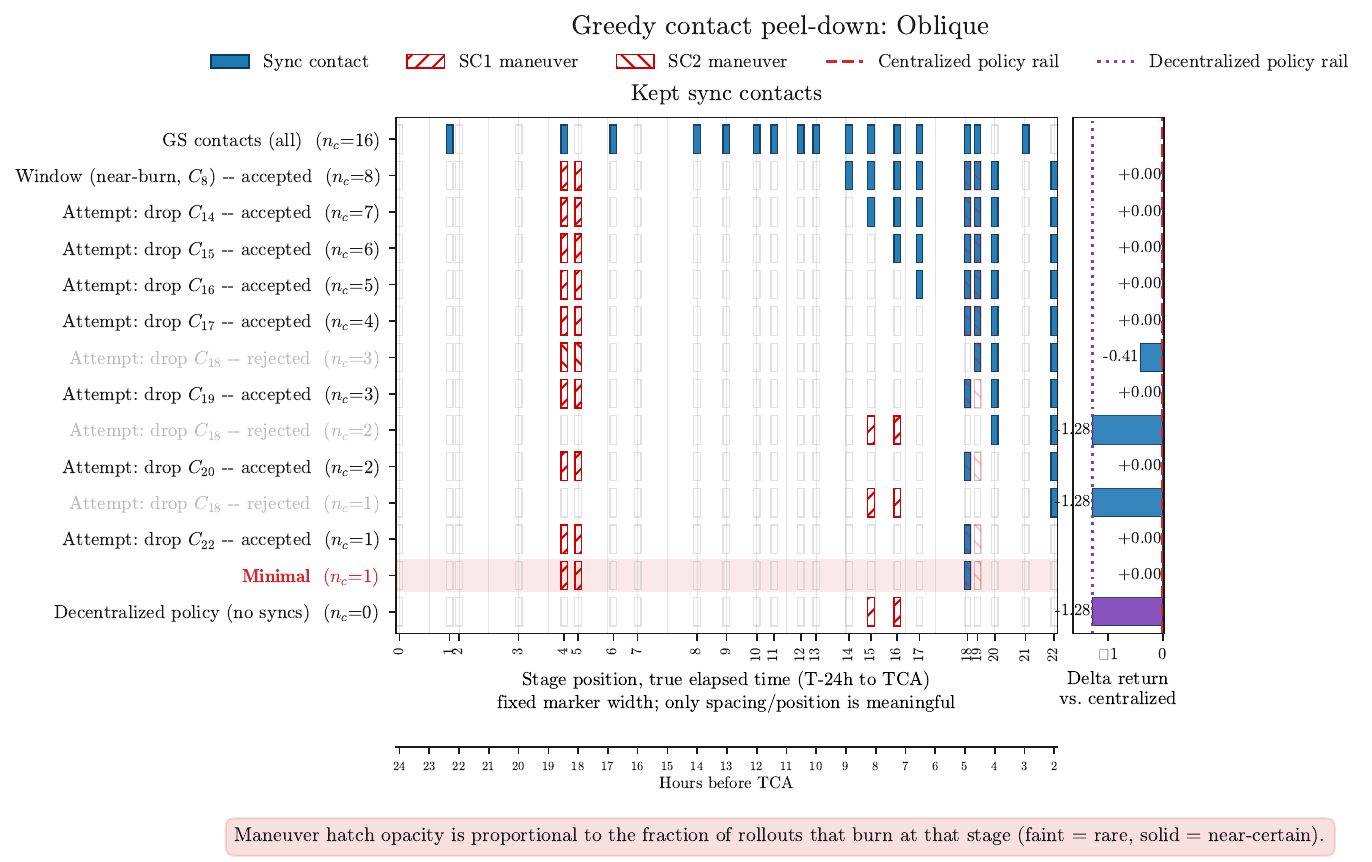}
    \caption{Greedy synchronization-schedule reduction for a representative oblique conjunction. Figure conventions follow Figure 8. Despite the different encounter geometry and contact schedule, the reduced solution again preserves synchronization near the dominant maneuver region while removing many earlier contacts.}
    \label{fig:peeloblique}
\end{figure}

\Cref{fig:peelheadon} presents the reduction process for a representative head-on conjunction. The initial synchronization schedule contains all communication opportunities generated from the ground-station network, after which synchronization events are progressively removed while preserving near-centralized performance. Several synchronization opportunities occurring well before maneuver execution are eliminated without measurable degradation in expected return. As the schedule becomes increasingly sparse, synchronization events immediately surrounding the maneuver are consistently retained, indicating that these communication opportunities contribute disproportionately to coordinated planning performance.

A similar trend is observed for the oblique conjunction shown in \Cref{fig:peeloblique}. Despite the different conjunction geometry and contact schedule, early synchronization opportunities are again preferentially removed while communication events near maneuver execution remain essential for maintaining solution quality. Together, these representative case studies suggest that synchronization timing, rather than synchronization frequency alone, governs the value of communication during collision avoidance planning, with opportunities surrounding maneuver execution contributing most to coordinated decision making.

%% file: Content/05_Conclusions.tex
\section{Conclusions}

This work presented a semi-decentralized planning framework for autonomous multi-spacecraft collision avoidance under realistic communication constraints. By deriving synchronization opportunities directly from predicted ground-station visibility windows, the proposed SDec-POMDP formulation explicitly models the asynchronous information exchange that characterizes current conjunction assessment operations, bridging the gap between idealized centralized planning and fully decentralized decision-making. The resulting planning problem is solved using approximate RS-SDA*, enabling coordinated maneuver planning under intermittent communication.

Across a representative suite of simulated conjunction scenarios, semi-decentralized planning consistently recovered near-centralized performance while requiring 28.5\% fewer synchronization events than continuous coordination. Further analysis demonstrated that communication opportunities surrounding maneuver execution contribute disproportionately to planning performance, indicating that synchronization timing is more important than synchronization frequency alone. Comparisons against representative rule-based operator heuristics further showed that communication-aware optimization not only maintains collision safety, but also produces maneuver strategies that more consistently satisfy operational mission objectives by avoiding unnecessary trajectory deviation and over-conservative collision avoidance. Future work will extend the framework to model ground-segment communication delays more explicitly and incorporate covariance-based probability-of-collision metrics alongside propagated miss distance.

More broadly, these results suggest a layered operational paradigm for autonomous conjunction assessment. In practice, conjunction assessment begins with a shared Conjunction Data Message (CDM), after which operators independently generate maneuver plans and receive updated spacecraft information as additional ground-station contacts become available. Under this workflow, an initial decentralized avoidance policy can provide a conservative safety baseline, while intermittent synchronization progressively refines maneuver decisions using updated state estimates and execution information. Rather than requiring continuous coordination, communication becomes an opportunity to improve maneuver quality whenever new information is available. By explicitly incorporating this communication structure into the planning process, the proposed semi-decentralized framework bridges the gap between idealized centralized planning assumptions and the operational realities of multi-operator collision avoidance, providing a practical foundation for future autonomous space traffic coordination.

%% file: Content/appendix.tex
\section*{Appendix A: Reduced State-Space Validation}
\label{app:state_validation}

The reduced planning state defined in ~\Cref{eqn:reduced_state} and ~\Cref{eqn:perp_distance} treats the perpendicular standoff $p_\perp$ as a scenario-specific constant and represents the terminal miss using ~\Cref{eqn:miss_distance}. The following validation examines the two assumptions underlying this reduction: that along-track control has negligible authority over $p_\perp$, and that the resulting frozen-$p_\perp$ representation reproduces the full three-dimensional miss at TCA.

\begin{figure}[H]
    \centering

    \begin{minipage}[c]{0.62\linewidth}
        \centering
        \includegraphics[width=\linewidth]{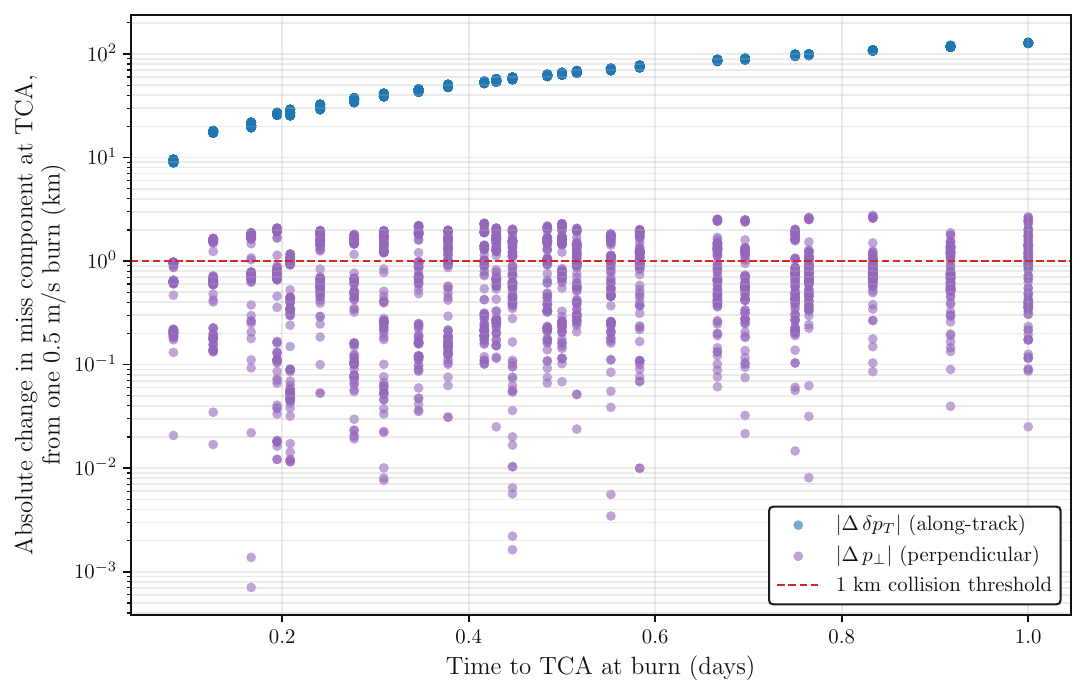}
    \end{minipage}
    \hfill
    \begin{minipage}[c]{0.35\linewidth}
        \centering
        \includegraphics[width=\linewidth]{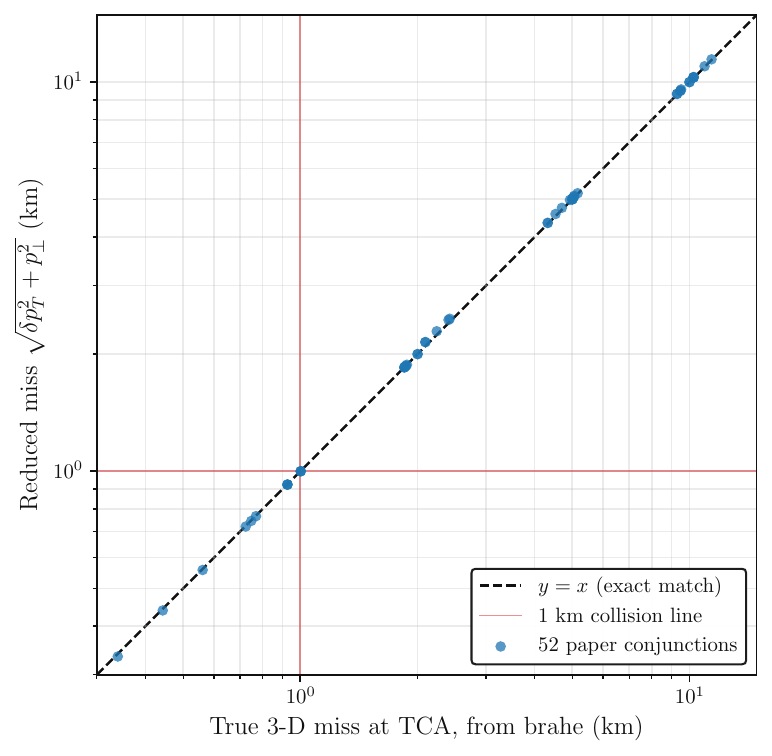}
    \end{minipage}

    \caption{Validation of the reduced conjunction-state representation.
    Left: a $0.5~\mathrm{m/s}$ along-track impulse changes $\delta p_T$ by tens of kilometers, while the induced change in $p_\perp=\sqrt{p_R^2+p_N^2}$ stays near the kilometer scale, arising only through higher-order coupling. Right: under ballistic propagation, the reduced
    miss $m=\sqrt{\delta p_T^2+p_\perp^2}$ with frozen $p_\perp$ reproduces
    the full three-dimensional Brahe miss at TCA across the 52-conjunction
    evaluation set, including near the $1~\mathrm{km}$ collision boundary.}
    \label{fig:appendix_state_validation}
\end{figure}

\emph{Approximation A: control has negligible authority over $p_\perp$.}
For each of the 52 conjunction geometries in the evaluation suite, a $0.5~\mathrm{m/s}$ along-track impulse of either sign was applied at each decision epoch and the resulting trajectory was propagated to TCA using the full six-dimensional Brahe model. As shown on the left of~\Cref{fig:appendix_state_validation}, the impulse changes $|\delta p_T|$ by roughly $10$--$130~\mathrm{km}$, while the induced change in $p_\perp$ stays at the kilometer scale or below across every geometry and burn time. %The median ratio $|\Delta p_\perp|/|\Delta\delta p_T|$ is $1.3\%$, and the maximum $|\Delta p_\perp|$ is $2.75~\mathrm{km}$. 
The available control therefore acts almost entirely along the retained state dimension; the residual perpendicular response is a higher-order effect, small relative to both the maneuver-induced along-track displacement and the state-discretization scale, so planning remains effective even with $p_\perp$ held fixed per conjunction.

\emph{Approximation B: the frozen-$p_\perp$ miss reproduces the true miss
at TCA.}
The ballistic reduced miss was evaluated for the same 52 conjunctions used
throughout the paper by combining the discretized $\delta p_T$ with
$p_\perp$ frozen at the encounter start. Independently, each conjunction
was propagated in the full six-dimensional Brahe environment and its true
three-dimensional miss was evaluated at TCA. Across true misses of
$0.07$--$11.4~\mathrm{km}$, the right of ~\Cref{fig:appendix_state_validation} lies on the
$y=x$ line, including near the $1~\mathrm{km}$ collision boundary. The
residual $m_{\mathrm{red}}-m_{\mathrm{3D}}$ has mean
$+0.017~\mathrm{km}$, mean absolute value $0.020~\mathrm{km}$, and maximum
absolute value $0.055~\mathrm{km}$; all 52 cases agree within
$\pm1~\mathrm{km}$. This comparison is explicitly ballistic and at TCA:
it isolates the fidelity of the state representation from the separate
maneuver-transition approximation.

Across the paper's complete conjunction suite, the reduced state therefore
preserves the terminal miss quantity used for decision making to well below
the discretization scale.

\section*{Appendix B: Operator-Heuristic Baselines}
\label{app:operator_heuristics}

The operator-heuristic baselines in ~\Cref{tab:operator_heuristics} outline various fixed, independent decision rules applied separately by each spacecraft. No heuristic constructs a coordinated joint maneuver plan. Each spacecraft maintains a continuous Gaussian estimate of its along-track miss and may execute an along-track avoidance burn followed by a return-to-nominal counter-burn (each $0.5~\mathrm{m/s}$). When a maneuver is selected, its direction and timing are chosen to target the center of the desired $4$--$7~\mathrm{km}$ miss-distance band, $m_{\mathrm{tgt}}=5.5~\mathrm{km}$.

\begin{table}[htbp]
\centering
\caption{Independent operator-heuristic baselines used for comparison.}
\label{tab:operator_heuristics}
\small
\setlength{\tabcolsep}{4pt}
\renewcommand{\arraystretch}{1.05}
\begin{tabular}{p{0.15\linewidth}p{0.53\linewidth}p{0.24\linewidth}}
\toprule
\textbf{Family} & \textbf{Rule and trigger} & \textbf{Purpose} \\
\midrule
Threshold &
Burn when the believed miss falls below $5~\mathrm{km}$ or $2~\mathrm{km}$. Execution is deferred to the latest stage from which one burn can reach $m_{\mathrm{tgt}}$; threshold $0$ denotes a nonmaneuvering spacecraft. &
Reactive go/no-go control. \\

Selfish &
Each spacecraft independently selects the latest burn predicted to place its own terminal miss near $m_{\mathrm{tgt}}$, using a blind, cautious, or observation-aware model of the other spacecraft. &
Smart but uncoordinated control. \\

Fixed-lead &
Execute one burn at $T-7~\mathrm{h}$ regardless of belief, then return toward the nominal trajectory. &
Schedule- versus belief-based timing. \\
\bottomrule
\end{tabular}
\end{table}